\documentclass[10pt]{article}
\usepackage[accepted]{tmlr}

\usepackage{graphicx}
\usepackage{psfrag}
\usepackage{amsmath,amsfonts}
\usepackage[inline,shortlabels]{enumitem}  \usepackage{wrapfig}
\usepackage{tikz}
\usepackage{lipsum}
\usepackage{nicefrac}
\usepackage{siunitx}
\usepackage{url}
\usepackage{pgfplots}
\usepackage{xspace}
\pgfplotsset{compat=newest}
\usepackage[shortcuts]{extdash}

\usepackage[pagebackref=true]{hyperref}
\renewcommand*\backref[1]{\ifx#1\relax \else ($^{\pmb\wedge}$\!~#1) \fi}

\usepackage{color, colortbl}
\definecolor{darkblue}{rgb}{0.0,0.0,0.65}
\definecolor{darkred}{rgb}{0.68,0.05,0.0}
\definecolor{darkgreen}{rgb}{0.0,0.29,0.29}
\definecolor{darkpurple}{rgb}{0.47,0.09,0.29}
\definecolor{myblue}{rgb}{0.1380392156862745, 0.3027450980392157, 0.6274509803921569}
\definecolor{myred}{rgb}{0.6235294117647059, 0.13725490196078433, 0.0196078431372549} 

\hypersetup{
  colorlinks = true,
  citecolor  = myblue,
  linkcolor  = myred,
  filecolor  = darkblue,
  urlcolor   = darkblue,
 }

\usepackage{xparse}
\ExplSyntaxOn
\NewDocumentCommand{\definealphabet}{mmmm}
 {\int_step_inline:nnn { `#3 } { `#4 }
   {
    \cs_new_protected:cpx { #1 \char_generate:nn { ##1 }{ 11 } }
     {
      \exp_not:N #2 { \char_generate:nn { ##1 } { 11 } }
     }
   }
 }
\ExplSyntaxOff

\definealphabet{v}{\mathbf}{a}{z}

\usepackage[nameinlink]{cleveref}
\crefname{equation}{Eq.}{Eqs.}
\Crefname{equation}{Equation}{Equations}
\crefname{table}{Tab.}{tables}
\Crefname{table}{Table}{tables}
\crefname{figure}{Fig.}{Figs.}
\Crefname{figure}{Figure}{figures}
\crefname{algorithm}{Alg.}{algorithms}
\Crefname{algorithm}{Algorithm}{algorithms}
\crefname{section}{Sec.}{sections}
\Crefname{section}{Section}{sections}
\Crefname{appendix}{Appendix}{Appendix}
\crefname{appendix}{App.}{Appendix}

\newcommand{\what}[1]{\widehat{#1}}
\newcommand{\uhat}{\what\vu}    \newcommand{\ourmodel}{Neural Split Operator }  \newcommand{\scinot}[2]{\texttt {#1}\mathrm{e}{{#2}}}  

\DeclareMathOperator{\fft}{\textsc{fft}}
\DeclareMathOperator{\ifft}{\textsc{ifft}}
\DeclareMathOperator{\physicsstep}{\operatorname{Physics-Step}}
\DeclareMathOperator{\nnstep}{\textsc{nn}}
\newcommand{\learnedcorrection}{\operatorname{Learned-Correction}}
\newcommand{\velocitysolve}{\operatorname{Velocity-Solve}}

\newcommand{\edit}[1]{{#1}}

\title{Learning to correct spectral methods \\
for simulating turbulent flows}

\author{\name Gideon Dresdner \email gideond@gmail.com \\
      \addr Google Research and ETH Zurich, Department for Computer Science
      \AND
      \name Dmitrii Kochkov \email dkochkov@google.com \\
      \addr Google Research
      \AND
      \name Peter Norgaard \email pnorgaard@google.com \\
      \addr Google Research
      \AND
      \name Leonardo Zepeda-N\'u\~nez \email lzepedanunez@google.com \\
      \addr Google Research
      \AND
      \name Jamie A. Smith \email jamieas@google.com \\
      \addr Google Research
      \AND
      \name Michael P. Brenner \email mbrenner@google.com \\
      \addr Google Research
      \AND
      \name Stephan Hoyer \email shoyer@google.com \\
      \addr Google Research
      }

\def\month{04}  \def\year{2023} \def\openreview{\url{https://openreview.net/forum?id=wNBARGxoJn}} 

\begin{document}

\maketitle

\begin{abstract}
Despite their ubiquity throughout science and engineering, only a handful of partial differential equations (PDEs) have analytical, or closed-form solutions. This motivates a vast amount of classical work on numerical simulation of PDEs and more recently, a whirlwind of research into data-driven techniques leveraging machine learning (ML). A recent line of work indicates that a hybrid of classical numerical techniques and machine learning can offer significant improvements over either approach alone. In this work, we show that the choice of the numerical scheme is crucial when incorporating physics-based priors. We build upon Fourier-based spectral methods, which are known to be more efficient than other numerical schemes for simulating PDEs with smooth and periodic solutions. Specifically, we develop ML-augmented spectral solvers for three common PDEs of fluid dynamics. Our models are more accurate ($\xspace2-4\times$) than standard spectral solvers at the same resolution but have longer overall runtimes ($\xspace{\sim 2\times}$), due to the additional runtime cost of the neural network component. We also demonstrate a handful of key design principles for combining machine learning and numerical methods for solving PDEs. \end{abstract}

\section{Introduction}
The numerical simulation of nonlinear partial differential equations (PDEs) permeates science and engineering, from the prediction of weather and climate to the design of engineering systems.
Unfortunately, solving PDEs on the fine grids required for high-fidelity simulations is often infeasible due to its prohibitive computational cost.
This leads to inevitable trade-offs between runtime and accuracy. The status quo is to solve PDEs on grids that are coarse enough to be computationally feasible but are often too coarse to resolve all phenomena of interest.
One classical approach is to derive coarse-grained surrogate PDEs such as Reynold's Averaged Navier Stokes (RANS) and Large Eddy Simulation (LES)~\citep{pope_turbulent_2000}, which in principle can be accurately solved on coarse grids.
This family of approaches has enjoyed widespread success but is tedious to perform, PDE-specific, and suffers from inherent accuracy limitations~\citep{Durbin2018-recent-developers-in-closure-modeling, pope_ten_2004}.

Machine learning (ML) has the potential to overcome many of these limitations by inferring coarse-grained models from high-resolution ground-truth simulation data. Turbulent fluid flow is an application domain that has already reaped some of these benefits. Pure ML methods have achieved impressive results, in terms of accuracy, on a diverse set of fluid flow problems  \citep{li_fourier_2021,stachenfeld_learned_2022}.
Going beyond accuracy, hybrid methods have combined classical numerical simulation with ML to improve stability and generalize to new physical systems, e.g., out of sample distributions with different forcing setups \citep{kochkov_machine_2021,list_learned_2022}.

However, hybrid methods have been limited to low-order finite-difference and finite volume methods, with the exception of one recent study~\citep{Frezat2022-fs}. Beyond finite differences and finite volumes, there is a broad field of established numerical methods for solving PDEs. In this paper, we focus on spectral methods which are used throughout computational physics \citep{trefethen_spectral_2000, Burns2020-gv} and constitute the core of the state-of-the-art weather forecasting system \citep{roberts2018climate}. Spectral methods, when applicable, are often preferred over other numerical methods because they can be more accurate for equations with smooth solutions. In fact, their accuracy rivals that of the recent progress made by ML. This begs the question: \emph{Can we improve spectral solvers of turbulent fluid flows using learned corrections of coarse-grained simulations?}

Our contributions are as follows:

\begin{enumerate}\item We propose a hybrid physics machine learning method that provides sub-grid corrections to classical spectral methods.
    \item We explore two toy 1D problems: the unstable Burgers' equation and the Kuramoto-Sivashinsky  the hybrid model is able to make the largest improvements.
    \item We compare spectral, finite-volume, ML-only, and our hybrid model on a 2D forced turbulence task. Our hybrid models provide some improvement on the accuracy of spectral-only methods, which themselves perform remarkably well compared to recently proposed ML methods. Furthermore, our results show that a key modeling choice for both hybrid and pure ML models is the use of velocity- rather than vorticity-based state representations.
\end{enumerate} 
\subsection{Related work}\label{sec:related_work}
The study of turbulent fluid dynamics is vast. We refer to \citet{pope_turbulent_2000} for a thorough introduction.
Classical approaches tend to derive mathematical approximations to the governing equations in an \emph{a priori} manner.
Recently, there has been an explosion of work in data-driven methods at the interface of computational fluid dynamics~(CFD) and machine learning~(ML). We loosely organize this recent work into three main categories:

\textbf{Purely Learned Surrogates}
fully replace numerical schemes from CFD with purely learned surrogate models. A number of different architectures have been explored, including multi-scale convolutional neural networks \citep{ronneberger_u-net_2015,wang_towards_2020}, graph neural networks \citep{sanchez-gonzalez_learning_2020}, and Encoder-Process-Decoder architectures \citep{stachenfeld_learned_2022}.

\textbf{Operator Learning}
seeks to learn the differential operator directly by mimicking the analytical properties of its class, such as pseudo-differential or Fourier integral operators but without explicit physics-informed components. These methods often leverage the Fourier transform \citep{li_fourier_2021,tran_factorized_2021} and the off-diagonal low-rank structure of the associated Green's function \citep{FanYing:mnnh2019,graph_fmm:2020}. 

\edit{
\textbf{Hybrid Physics-ML}
is an emerging set of approaches which aim to combine classical numerical methods with contemporary data-driven deep learning techniques.
These approaches use high-resolution simulation data to learn corrections to low-resolution numerical schemes with the goal of combining the best of both worlds --- the simplicity of PDE-based governing equations and the expressive power of neural networks. Our work falls into this category, along with a growing body of work \citep{mishra_machine_2019,bar-sinai_learning_2019,kochkov_machine_2021,Bruno:2021_fc_dyn_nn_shocks,list_learned_2022,Frezat2022-fs,huang_accelerating_2022}.

Some of these works stand out as being closely related to ours but with important differences. \citet{huang_accelerating_2022} also develops a hybrid approach but focuses exclusively on ODEs, not PDEs, and it is not clear whether their techniques for stable simulations with large time-step size would also apply to PDEs. In their recent work, \citet{Frezat2022-fs} augmented spectral solvers for closure models exclusively focused on climate modeling. We also learn corrections to spectral methods but rather than focusing on climate models, we tackle more general PDEs, provide key ML architecture choices, and include comparisons to classical numerical methods on multiple grid resolutions.

Unlike our work, \citet{san_extreme_2018} and \citet{subel_data-driven_2021} take fundamentally different approaches to ours. \citet{san_extreme_2018} improves classical approaches for closure modeling by using a neural network with a single hidden layer to learn fine-tuned local corrections instead of the usual global correction methods. Their work requires specialized knowledge of classical methods (POD, closure models, etc.) which is orthogonal to our larger goal of automating these approaches using data-driven techniques. \citet{subel_data-driven_2021} focus on a data augmentation scheme to overcome stationary shocks that arise in Burgers’ equation. Their proposed shifting mechanism could be used in conjunction with our work, but represents a different research direction.
} 
\section{Spectral methods for fluids}

Spectral methods are a powerful method for finding high-accuracy solutions to PDEs and
are often the method of choice for solving smooth PDEs with simple boundary conditions and geometries.
There is an extensive literature on the theoretical and practical underpinnings of spectral methods~\citep{gottlied_orszag1977spectral,gottlieb_spectral_2001,canuto:2007spectral,Implementing_SM_PDEs:2009}, particularly for methods based on Fourier spectral collocation \citep{trefethen_spectral_2000,Boyd:01_Chebyshev_Fourier}. Below, we provide a succinct introduction.

\subsection{Partial differential equations for turbulent fluids}

\begin{samepage}
Let $\vu: \mathbb{R}^d \times \mathbb{R}^{+} \rightarrow \mathbb{R}^{d'}$ be a time-varying vector field, for dimensions $d$ and $d'$. We study PDEs of the form
\begin{equation}\label{eq:pde_def}
    \partial_t \vu = D\vu + N(\vu),
\end{equation}
plus initial and boundary conditions. Here $D$ is a linear partial differential operator, and $N$ is a non-linear term.
\end{samepage}
Equations of this form dictate the temporal evolution of $\vu$ driven by its variation in space. 
In practice, PDEs are solved by discretizing in space and time, which converts the continuous PDE into a set of update rules for vectors of coefficients to approximate the state $\vu$ in some discrete basis, e.g., on a grid.
For time-dependent PDEs, temporal resolution must be scaled proportionally to spatial resolution to maintain an accurate solution. Thus, runtime for PDE solvers is $O(n^{d+1})$, where $d$ is the number of spatial dimensions and $n$ is the number of discretization points per dimension.

For simulations of fluids, the differential operator is typically either diffusive, $D = \partial_x^{2}$, or hyper-diffusive, $D = \partial_x^{4}$. And, the non-linearity is a convective term, $N = \frac{1}{2} \partial_x (\vu^2) = \vu\,\partial_x \vu$.
\emph{Diffusion} is the tendency of the fluid velocity to become uniform due to internal friction, and \emph{convection} is the tendency of the fluid to be transported by its own inertia.
Turbulent flows are characterized by a convective term that is much stronger than diffusion.

For most turbulent flows of interest, closely approximating the exact PDE (known as Direct Numerical Simulation) is computationally intractable because it requires prohibitively high grid resolution.
Instead, coarse-grained approximations of the PDE are solved, known as ``Large Eddy Simulation'' (LES).
LES augments \Cref{eq:pde_def} by adding a correction term determined by a closure model to account for averaged effects over fine spatial length scales.
In practice, this term is often omitted due to the difficulty of deriving appropriate closure formulas (``implicit LES'') and the PDE is simply simulated with the finest computationally feasible grid resolution.
In our case, correction terms are an opportunity for machine learning.
If we can accurately model PDEs on coarser grids with suitable correction terms, we may be able to significantly reduce the computational cost of large-scale simulations.

\subsection{The appeal of the Fourier basis for modeling PDEs}\label{sec:fourier_basis}

\begin{figure}[t!]
\vspace{-7mm}
    \includegraphics[width=\textwidth]{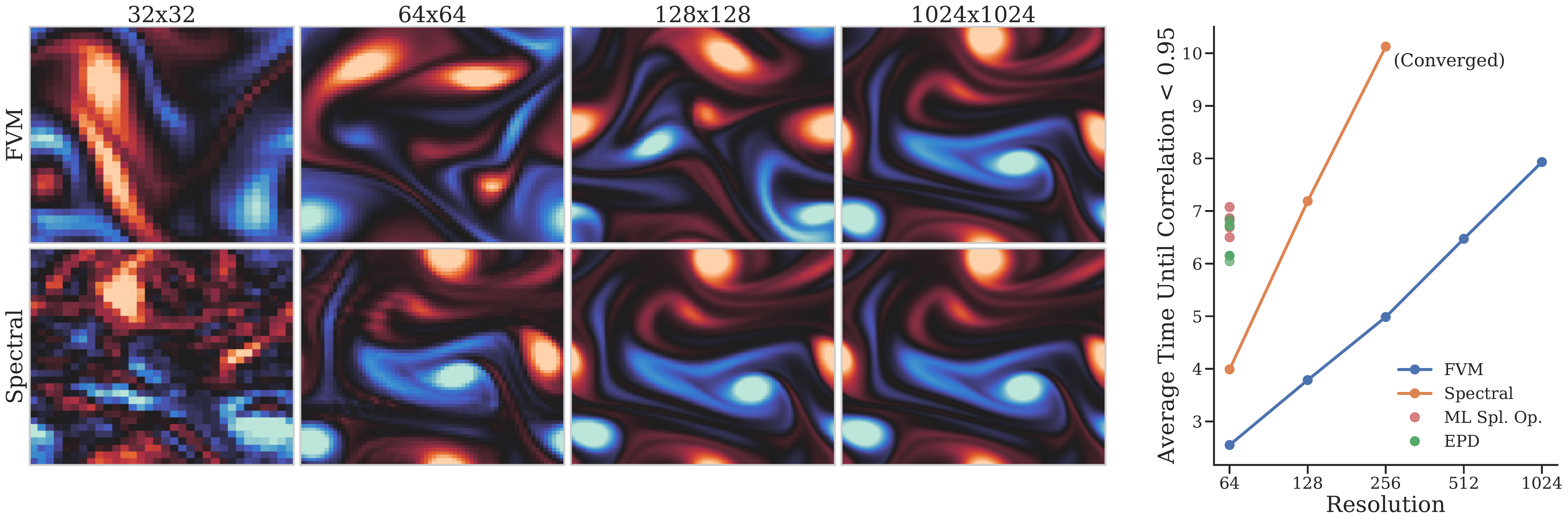}
    \caption{Comparing finite volume (FVM) to spectral convergence for 2D turbulence. \emph{Left}: Vorticity fields evolved up to time 9.0 with different grid sizes and numerical methods, starting from an identical initial condition. Qualitatively, it is clear that at resolution 64x64, the spectral method has already captured most of features of the high-resolution state. The FVM looks sharp, but clearly differs. At sufficiently high resolution, the methods converge to the same solution. \emph{Right}: Here we compare each method with high-resolution 2048x2048 ground truth. We plot the time until the first dip below 0.95 of the correlation with the ground truth. Note that the initial conditions for the finite volume and the spectral method are sampled from the same distribution, but are not identical. At resolution 256, the spectral method has converged within the accuracy limitations of single precision, so we omit higher resolutions. For the EPD baseline and our ML Split Operator (ML Spl.\ Op.) models, we show performance across five different neural network parameter initializations.
    }
    \label{fig:fvm_vs_spectral}
\end{figure}

Let us further assume that $\vu(x,t):[0, 2\pi]\times\mathbb R^+\to\mathbb R$ in \Cref{eq:pde_def} is $2\pi$-periodic, square-integrable for all times~$t$, and for simplicity, one-dimensional.
Consider the Fourier coefficients $\uhat^t$ of $\vu(x,t)$, truncated to lowest $K+1$ frequencies (for even $K$):
\begin{equation}\label{eq:spectral_basis}
\uhat^t = (\uhat^t_{-\nicefrac{K}{2}}, \ldots, \uhat_k^t, \ldots,\uhat^t_{\nicefrac{K}{2}}) \quad\text{where}\quad \what\vu_k^t = \frac{1}{2 \pi}\int_0^{2\pi} \vu(x,t) e^{-ik\cdot x}dx.
\end{equation}
$\uhat^t$ is a vector representing the solution at time $t$, containing the coefficients of the Fourier basis $e^{ikx}$ for $k\in\{-\nicefrac{K}{2},\ldots,\nicefrac{K}{2}\}$. In general, the integral $\uhat^t_k$ has no analytical solution and so we approximate it using a trapezoidal quadrature on $K+1$ points. Namely, we approximate it by sampling $\vu(x,t)$ on an equispaced grid. The Fast Fourier Transform (FFT) \citep{Cooley_Tukey:1965} computes these Fourier coefficients efficiently in log-linear time.
Spectral methods for PDEs leverage the fact that differentiation in the Fourier domain can be calculated by element-wise multiplication according to the identity $\partial_x \uhat_k = i k \uhat_k$.
This, in turn, makes inverting linear differential operators easy since it is simply element-wise division~\citep{trefethen_spectral_2000}.

When time-dependent PDEs include non-linear terms, spectral methods evaluate these terms on a grid in real-space, which requires forward and inverse FFTs inside each time-step.
These transforms introduce two sources of error. First, the quadrature rules for $\uhat^t_k$ produce only an approximation to the integral. 
Second, there will be a truncation error when the number of frequencies $K+1$ is less than the bandwidth of $\vu(x,t)$. Remarkably, it is a well-known fact of Fourier analysis that both of these errors decay super-algebraically (i.e., faster than any power of $1/K$)
if $\vu(x,t)$ is periodic and is in $C^\infty$ \citep{trefethen_spectral_2000}.\footnote{
Convergence is exponential for analytic $\vu$ \citep{Tadmor:1986_exp_acc}.
} Thus, relatively few discretization points are needed to represent solutions which are $C^\infty$.

Because of these favorable convergence properties, spectral methods often outshine their finite difference counterparts.
For example, spectral methods are used for large-scale simulations of turbulence on GPU super-computers \citep{Yeung2020-3d-turbulence}.
See \Cref{fig:fvm_vs_spectral} for a simple comparison in the case of 2D turbulence. This motivates us to start from spectral methods, rather than finite difference methods, and further improve them using machine learning.
 
\section{Learned split operators for correcting spectral methods}
Similarly to classical spectral numerical methods, we solve \Cref{eq:pde_def} by representing our state in a finite Fourier basis as $\uhat^t$ and integrating forward in time.
We model the fully known differential operators $D$ and~$N$ using a standard spectral method, denoted $\physicsstep$.
The machine learning component, denoted $\learnedcorrection(\,\cdot\,; \theta)$, contains tune-able parameters $\theta$ which are optimized during training to match high-resolution simulations.

Building upon traditional physics-based solvers,
we use a simple explicit-Euler time integrator for the correction term \edit{(See \cref{sec:cfd} for a brief explanation of explicit methods.)} This yields the following update equation:
\begin{equation}\label{eq:our_model}
    \uhat^{t+1} = \physicsstep(\uhat^t) + h\cdot \learnedcorrection(\uhat^t; \theta),
\end{equation}
where $h\in\mathbb R$ is the time step size.
\edit{Ultimately, our work is aimed at two and three dimensional problems so rather than present an abstract schematic as presented in the equation above (\cref{eq:our_model}), we chose to present a more detailed, realistic schema of 2D turbulence (\cref{eq:navier_stokes}). The spatial state variable is thus not the generalized $\vu$ as in \Cref{eq:our_model} but rather, the vorticity $\omega$ as defined in \Cref{sec:navier_stokes}}

\subsection{Convolutional layers for encoding the physical prior of locality}\label{sec:locality_ansatz}

For a small time step~$h$, the solution at time $t+h$ only depends locally in space on the solution at time $t$ (within a dependency cone usually encoded by the Courant–Friedrichs–Lewy (CFL) condition). 
Following previous work (see \cref{sec:related_work}, Hybrid Physics-ML), we incorporate this assumption into the machine learning component of our model using Convolutional Neural Networks (CNN or ConvNet) which,  by design, only learn local features. We now provide a high-level view of our modeling choices:

\paragraph{Real-space vs.\ frequency-space.} Since the Fourier basis is global, each coefficient $\uhat_k^t$ contains information from the full spatial domain as shown in \Cref{eq:spectral_basis}. Thus, in order to maintain spatial features, which are local, we apply the ConvNet component of our \ourmodel model in real-space. This is accomplished efficiently via inverse-FFT and FFT to map the signal back and forth between frequency- and real-space.

\paragraph{ConvNet Padding.} In this work, we tackle problems with periodic boundary conditions. This makes periodic padding a natural choice for the ConvNets.

\paragraph{Neural architecture.} \edit{For both 1D and 2D problems we used an Encoder-Process-Decoder (EPD) architecture~\citep{stachenfeld_learned_2022} to facilitate fairer comparisons to pure neural network baselines. While various forms of optimization is possible, such as model parallelism among others, we consider this to be outside the scope of this exploratory work.} See \Cref{sec:baselines} for a detailed description of EPD models and \Cref{app:architectures} for further information.

\subsection{The split operator method for combining time scales}

\begin{figure}
    \centering
\fontsize{12pt}{14.4pt}\selectfont \def\svgwidth{.9\textwidth}
    \begingroup \makeatletter \providecommand\color[2][]{\errmessage{(Inkscape) Color is used for the text in Inkscape, but the package 'color.sty' is not loaded}\renewcommand\color[2][]{}}\providecommand\transparent[1]{\errmessage{(Inkscape) Transparency is used (non-zero) for the text in Inkscape, but the package 'transparent.sty' is not loaded}\renewcommand\transparent[1]{}}\providecommand\rotatebox[2]{#2}\newcommand*\fsize{\dimexpr\f@size pt\relax}\newcommand*\lineheight[1]{\fontsize{\fsize}{#1\fsize}\selectfont}\ifx\svgwidth\undefined \setlength{\unitlength}{1114.01574803bp}\ifx\svgscale\undefined \relax \else \setlength{\unitlength}{\unitlength * \real{\svgscale}}\fi \else \setlength{\unitlength}{\svgwidth}\fi \global\let\svgwidth\undefined \global\let\svgscale\undefined \makeatother \begin{picture}(1,0.26414134)\lineheight{1}\setlength\tabcolsep{0pt}\put(0,0){\includegraphics[width=\unitlength,page=1]{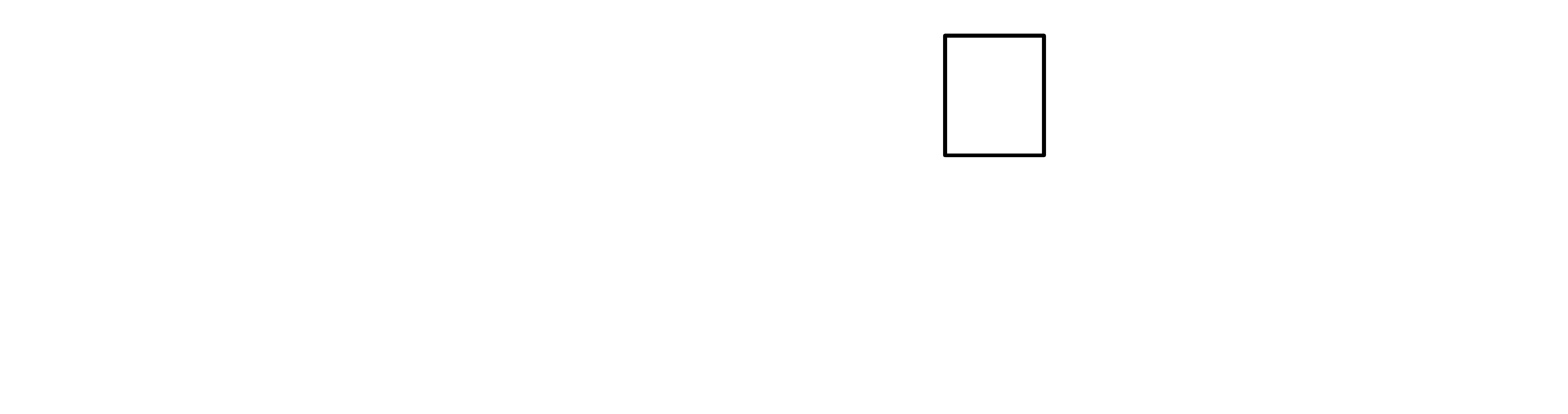}}\put(0.00033613,0.11869971){\color[rgb]{0,0,0}\makebox(0,0)[lt]{\lineheight{1.25}\smash{\begin{tabular}[t]{l}$\what\omega^t$\end{tabular}}}}\put(0,0){\includegraphics[width=\unitlength,page=2]{inkscape/drawing.pdf}}\put(0.20165052,0.04233686){\color[rgb]{0,0,0}\makebox(0,0)[lt]{\lineheight{1.25}\smash{\begin{tabular}[t]{l}$\physicsstep$\end{tabular}}}}\put(0,0){\includegraphics[width=\unitlength,page=3]{inkscape/drawing.pdf}}\put(0.52253416,0.04020909){\color[rgb]{0,0,0}\makebox(0,0)[lt]{\lineheight{1.25}\smash{\begin{tabular}[t]{l}$\what\omega^{t+1}_{\physicsstep}$\end{tabular}}}}\put(0,0){\includegraphics[width=\unitlength,page=4]{inkscape/drawing.pdf}}\put(0.919807,0.19724367){\color[rgb]{0,0,0}\makebox(0,0)[lt]{\lineheight{1.25}\smash{\begin{tabular}[t]{l}$\cdot h$\end{tabular}}}}\put(0.44528583,0.20092785){\color[rgb]{0,0,0}\makebox(0,0)[lt]{\lineheight{1.25}\smash{\begin{tabular}[t]{l}$(\vv_x^t, \vv_y^t)$\end{tabular}}}}\put(0,0){\includegraphics[width=\unitlength,page=5]{inkscape/drawing.pdf}}\put(0.31593308,0.19948319){\color[rgb]{0,0,0}\makebox(0,0)[lt]{\lineheight{1.25}\smash{\begin{tabular}[t]{l}$\ifft$\end{tabular}}}}\put(0,0){\includegraphics[width=\unitlength,page=6]{inkscape/drawing.pdf}}\put(0.7399705,0.19927816){\color[rgb]{0,0,0}\makebox(0,0)[lt]{\lineheight{1.25}\smash{\begin{tabular}[t]{l}$\fft$\end{tabular}}}}\put(0,0){\includegraphics[width=\unitlength,page=7]{inkscape/drawing.pdf}}\put(0.92179141,0.10545171){\color[rgb]{0,0,0}\makebox(0,0)[lt]{\lineheight{1.25}\smash{\begin{tabular}[t]{l}$+$\end{tabular}}}}\put(0,0){\includegraphics[width=\unitlength,page=8]{inkscape/drawing.pdf}}\put(1.00469798,0.10718226){\color[rgb]{0,0,0}\makebox(0,0)[lt]{\lineheight{1.25}\smash{\begin{tabular}[t]{l}$\what\omega^{t+1}$\end{tabular}}}}\put(0,0){\includegraphics[width=\unitlength,page=9]{inkscape/drawing.pdf}}\put(0.04792013,0.19993646){\color[rgb]{0,0,0}\makebox(0,0)[lt]{\lineheight{1.25}\smash{\begin{tabular}[t]{l}$\velocitysolve$\end{tabular}}}}\put(0,0){\includegraphics[width=\unitlength,page=10]{inkscape/drawing.pdf}}\put(0.61689788,0.19724346){\color[rgb]{0,0,0}\makebox(0,0)[lt]{\lineheight{1.25}\smash{\begin{tabular}[t]{l}$\nnstep$\end{tabular}}}}\put(0,0){\includegraphics[width=\unitlength,page=11]{inkscape/drawing.pdf}}\put(0.51659657,0.25633096){\color[rgb]{0,0,0}\makebox(0,0)[lt]{\lineheight{1.25}\smash{\begin{tabular}[t]{l}$\learnedcorrection$\end{tabular}}}}\end{picture}\endgroup  \caption{Diagram summarizing our model described in \Cref{eq:our_model} for the 2D Navier-Stokes equation. The input, vorticity $\widehat\omega^t$, is processed by two independent components\---$\learnedcorrection$ and $\physicsstep$\---operating at different time-scales. The output of $\learnedcorrection$ is weighted by $h$, as in a basic first-order Euler stepping scheme, and combined with the output of the $\physicsstep$ to give the state at the next time step. 
    }
    \label{fig:model-diagram}
\end{figure}

Due to a variety of considerations --- numerical stability, computational feasibility, etc. --- each term of a PDE often warrants its own time-advancing method.
This motivates \emph{split operator methods}~\citep{Strang:1968_operator_splitting,mclachlan_quispel_2002}, a popular tool for solving PDEs which combine different time integrators. In this work, we use the split operator approach to incorporate the additional terms given by the neural network.

In the usual pattern of spectral methods, $\physicsstep$ itself is split into two components corresponding to the $D$ and $N$ terms of \Cref{eq:pde_def}.
The $D$ term of $\physicsstep$ is solved with a Crank-Nicolson method and the $N$ part is solved using explicit 4th-order Runge-Kutta, which effectively runs at a time-step of $h/4$.
The $\learnedcorrection$ component is solved using a vanilla first-order Euler time-step with step size $h$, which when compared to incorporating the learned component in the 4th-order Runge-Kutta solver, is more accurate and has 4x faster runtime (see \cref{sec:results}).
Alternatively, omitting $\physicsstep$ from \Cref{eq:our_model} gives a Neural ODE model \citep{chen_neural_2019} with first-order time-stepping which is precisely the EPD model described by \citet{stachenfeld_learned_2022}. This model serves as a strong baseline as shown in \Cref{sec:results}.

\subsection{Physics-based solvers for calculating neural-net inputs}

A final important consideration is the choice of input representation for the machine learning model. We found that ML models operate better in velocity-space whereas vorticity is the more suitable representation for the numerical solver. We were able to improve accuracy by incorporating a physics-based data pre-processing step, e.g., a velocity-solve operation, for the inputs of the neural network component \edit{(See \cref{sec:cfd} for a brief explanation of velocity-solve)}.
Our overall model is $\learnedcorrection = \fft(\nnstep(\ifft(\operatorname{State-Transform}(\uhat^t))))$, where $\nnstep$ is implemented a periodic ConvNet.
For our 1D test problems, State-Transform is the identity transformation, but for 2D Navier-Stokes (depicted in \cref{fig:model-diagram}) we perform a velocity-solve operation to calculate velocity from vorticity.
This turns out to be key modeling choice, as described in \Cref{sec:results}.

\subsection{Training and evaluation}\label{sec:training_setup}

\paragraph{Data preparation for spectral solvers.}

For ground truth data, we use fully resolved simulations, which we then downsample to the coarse target resolution. Choosing the downsampling procedure is a key decision for coarse-grained solvers~\citep{Frezat2022-fs}. In this work, the fully resolved trajectories are first truncated to the target wavenumber (i.e., an ideal low-pass filter). Then, we apply an exponential filter of the form $\tilde\vu_k = \exp( -\alpha | k / k_\text{max} |^{2p}) \uhat_k$, where $\tilde\vu_k$ denotes the filtered field and $k$ is the $k$-th wavenumber~\citep{Canuto2006-xw}.
We obtained stable trajectories using a relatively weak filter with $\alpha = 6$ and $p = 16$.
The exponential filter smooths discontinuities in the PDE solution, which otherwise manifest themselves globally in real space as ringing artifacts known as Gibbs Phenomena~\citep{Canuto2006-xw}.

Filtering is also used to correct aliasing errors in spectral methods that arise when evaluating non-linear terms~\citep{gottlieb_spectral_2001}.
Ideally, filtering for aliasing errors would be spatially adaptive~\citep{Boyd1996-wy}.
In practice, however, filters for both aliasing and truncation are often chosen heuristically.
While one might aspire to learn these heuristics from data, our attempts at doing so were unsuccessful. In part this is because insufficiently filtered simulations will often entirely diverge rather than accumulate small errors.
Thus, in addition to filtering the downsampled training data, we also used the exponential filter on the outputs of $\physicsstep$ of \Cref{eq:our_model} and consider this to be another component of $\physicsstep$.
Omitting this filtering also resulted in global errors which were impossible for the learned component to correct.
\Cref{fig:training_diagram} summarizes our data generation pipeline, including an explicit filtering step.
In \Cref{sec:results}, we present a failure mode on a model without filtering (see \cref{fig:unstable_burgers}).

\edit{We experimented with larger time-step sizes, e.g. $h=0.1$, $h=1.0$ but found the resulting models to be unstable. This is because the physics-component of the model quickly becomes unstable at large step sizes which propagates to the overall model.}

\paragraph{Training loss.}
We train our models to minimize the squared-error over some number of unrolled prediction steps, which allows our model to account for compounding errors \citep{um_solver---loop_2021}.
Let $\overline\vu(x,t)$ denote our prediction at time $t$, then our training objective is given by $ \beta \sum_{x,t\leq T} |\overline\vu(x,t) - \vu(x,t)|^2$. The scaling constant $\beta$ is chosen so that the loss of predicting ``no change'' is one, i.e., $\beta^{-1} = \sum_{x,t\leq T} |\vu(x,0) - \vu(x,t)|^2$.
Relative to the thousands of time-steps over which we hope to simulate accurately, we unroll over a relatively small number during training (e.g., $T=32$ for 2D turbulence) because training over long time windows is less efficient and less stable~\citep{kochkov_machine_2021}.

\paragraph{Measuring convergence.}
We seek to optimize the accuracy of coarse-grained simulations. More specifically, we aim to make a coarse resolution simulation as similar as possible to a high-resolution simulation which has been coarse-grained in post-processing. Following \citet{kochkov_machine_2021}, we measure convergence to a fully resolved, high-resolution ground-truth at each time $t$ in terms of mean absolute error (MAE), correlation, and time until correlation is less than \num{0.95}.
MAE is defined as $\sum_x |\overline\vu(x,t) - \vu(x,t)|$ and correlation is defined $\operatorname{Corr}[\vu(\cdot, t), \overline{\vu}(\cdot, t)] = \sum_x (\overline\vu(x,t) \cdot \vu(x,t)) / (\|\overline\vu(x,t)\|_2 \|\vu(x,t)\|_2)$ (since our flows have mean zero). Finally, we compute the first time step in which the correlation dips below \num{0.95}, i.e.\ \mbox{$\min \{t\mid \operatorname{Corr}[{\vu}(\cdot, t),  \overline{\vu}(\cdot, t)] < 0.95\}$}. While MAE is precise up to floating point precision, it is not readily interpretable. Whereas, correlation is less sensitive but more interpretable. For this reason, we prefer to report correlation. This potential redundancy is demonstrated in our experiment on the KS equation~(\cref{sec:ks}), where we included both MAE and correlation.

For the 2D Navier-Stokes equation, we only report correlation because it is sufficient. For the unstable Burgers’ equation, the situation was the opposite: We only reported MAE because measuring correlation did not provide additional insight — all models had correlation values close to \num{1.0}, whereas MAE was sensitive to the improved performance of our method.
 
\section{Results}\label{sec:results}
\begin{figure}[t!]
    \centering
\includegraphics[width=.8\textwidth]{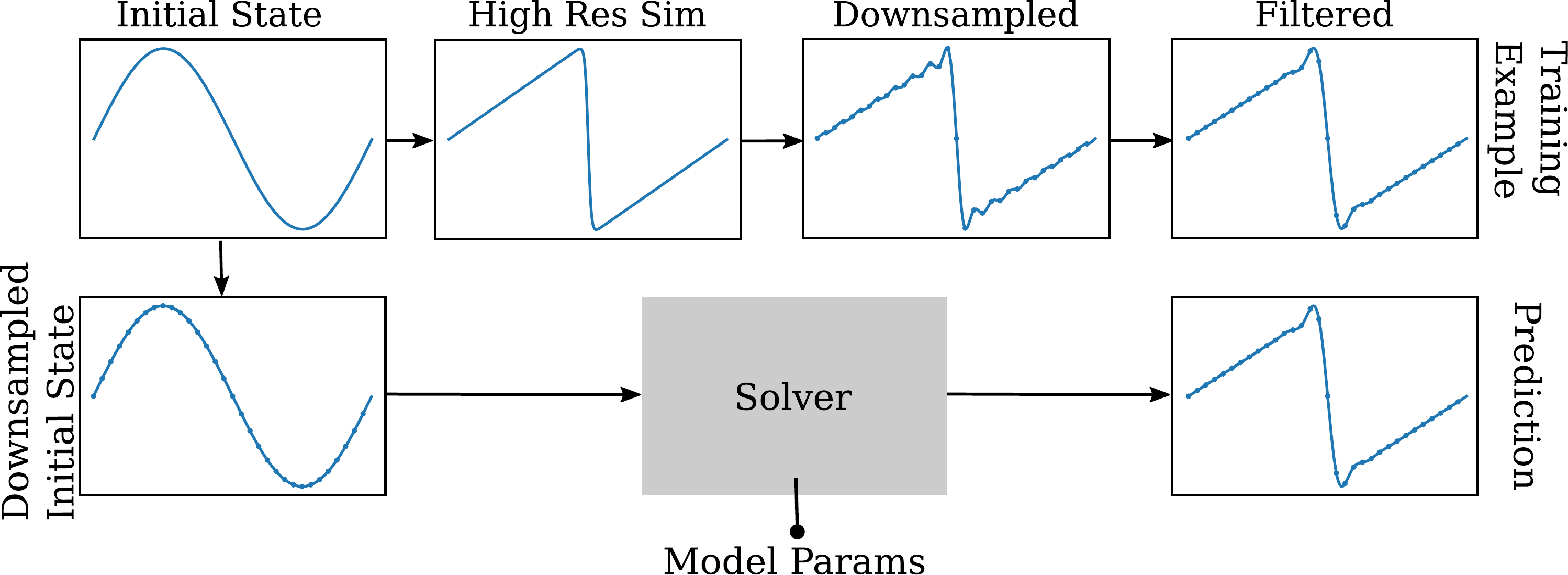}
    \caption{Diagram of our training pipeline. Starting with a high resolution initial state, we run it forward using a high-resolution spectral solver. Then we downsample by truncating higher frequencies in the Fourier representation. This can cause ``ringing effects'' for which the standard approach is to apply a filtering operator. Finally, we train a solver to mimic this process as closely as possible, as measured in $\ell_2$-loss.}
    \label{fig:training_diagram}
\end{figure}

\subsection{Model equations}\label{sec:model_equations}

We showcase our method using three model equations which capture many of the algorithmic difficulties present in 
more complex systems: 1D Kuramoto–Sivashinsky (KS) equation, 1D unstable Burgers' equation, and 2D Kolmogorov flow, a variant of incompressible Navier-Stokes with constant forcing.

The KS equation has smooth solutions which spectral methods are well-designed to solve. Therefore, it is not surprising that, while our method does improve over spectral-only methods, that improvement is not significant.
On the other hand, the unstable Burgers' equation presents a test case in which classical spectral methods struggle near discontinuities. Here, our method outperforms spectral-only methods. Finally, with two-dimensional Kolmogorov flow, we demonstrate our method on a more challenging fluid simulation. Kolmogorov flow exhibits multiscale behavior in addition to smooth, chaotic dynamics. Without any modification, spectral-only methods are already competitive with the latest hybrid methods \citep{kochkov_machine_2021}. Similar to the 1D KS equation, our method is able to provide some improvement in this case.

\subsection{Baselines}\label{sec:baselines}

Our hybrid spectral-ML method naturally gives rise to two types of baselines: spectral-only and ML-only. On all three equations, we compare to spectral-only at various resolutions, e.g., ``Spectral 32'' refers to the spectral-only method with 32 grid points.

For Kolmogorov flow, we compare to two additional baselines:
\begin{enumerate*}[(1)]
    \item the Encoder-Process-Decoder~(EPD) model \citep{stachenfeld_learned_2022}, a state-of-the-art ML-only model, and
    \item a finite volume method~(FVM) as implemented by \cite{kochkov_machine_2021}~---~a standard numerical technique which serves as an alternative to the spectral method
\end{enumerate*}.

The EPD architecture acts in three steps. First, the spatial state is embedded into a high-dimensional space using a feed-forward neural network encoder architecture. Then, either another feed-forward or a recurrent architecture is applied to process the embedded state. Finally, the output of the process step is projected to back to the spatial state using another feed-forward decoder module. This method has recently achieved state-of-the-art on a wide range of one-, two-, and three-dimensional problems. For consistency, we used an identical EPD model for the learned component of our model (\cref{eq:our_model}).

Finally, we included a second-order FVM on a staggered grid to illustrate the strength of the spectral baselines. This gives context to compare to previous work in hybrid methods \citep{bar-sinai_learning_2019,kochkov_machine_2021} which use this FVM model as a baseline.

We avoided extensive comparisons to classical subgrid modeling, such as Large Eddy Simulation (LES), since LES is itself a nuanced class of methods with many tunable parameters. Instead, our physics-only baselines are implicit LES models --- a widely-used and well-understood model class which serves as a consistent, parameter-free baseline. In contrast, explicit sub-grid-scale models such as Smagorinsky models include tunable parameters, with the optimal choice depending on the scenario of interest. Furthermore, explicit LES models typically focus on matching the energy spectrum rather than minimizing point-wise errors \citep{list_learned_2022}.

\subsection{Kuramoto-Sivashinsky (KS) equation}\label{sec:ks}
 The Kuramoto-Sivashinsky (KS) equation is a model of unstable flame fronts.
In the PDE literature, it is a popular model system because it exhibits chaotic dynamics in only a single dimension.
The KS equation is defined as
\begin{equation}\label{eq:ks_eq}
    \partial_t \vu = -\vu \partial_x \vu - \partial_x^4 \vu - \partial_x^2 \vu.
\end{equation}
The three terms on the right hand side of this equation correspond to convection, hyper-diffusion and anti-diffusion, which drives the system away from equilibrium.

Because solutions are smooth, spectral methods are able to capture the dynamics of the KS equation extremely well. At a resolution of 64, the simulation is already effectively converged to the limits of single precision arithmetic, i.e., it enjoys a perfect correlation with a high-resolution ground truth.

Considering how well-suited spectral methods are for modeling this equation, it is remarkable that our ML-Physics hybrid model is able to achieve any improvement at all. Looking qualitatively at the left panel of \Cref{fig:ks}, there is a clear improvement over the spectral 32 baseline.

\subsection{Unstable Burgers' equation}\label{sec:unstable_burgers}

\begin{figure}[t!]
    \includegraphics[width=1.0\textwidth]{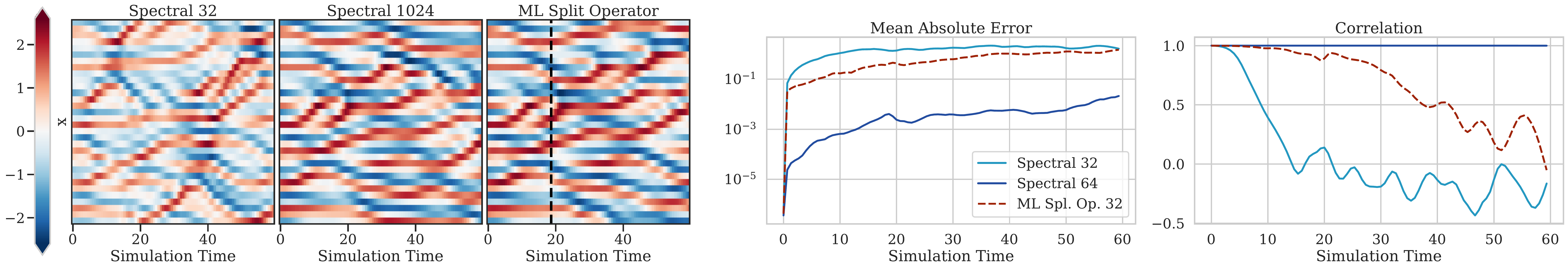}
    \caption{Comparing our model to spectral-only baselines on the KS equation. The spectral method is essentially converged to a 1024 ground truth model at a resolution of 64. Vertical dashed line indicates the first time step in which our model's correlation with the ground truth is less than 0.9.
    Our model is still able to some improvement over a coarse resolution (Spectral 32) baseline.
    }
    \label{fig:ks}
    \bigskip
    \includegraphics[width=1.0\textwidth]{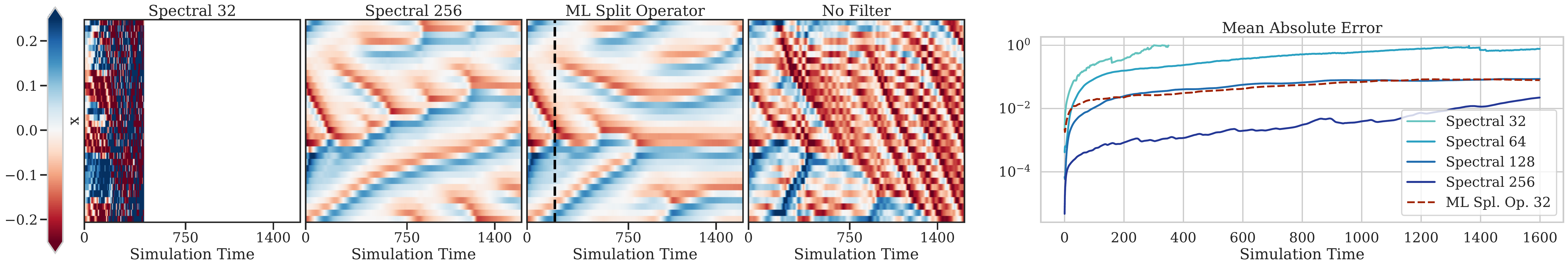}
\caption{Comparing our model to spectral-only baselines on the unstable Burgers'\ equation~(\cref{eq:unstable_burgers}). 
    Vertical dashed line indicates the first time step in which our model's correlation with the ground truth is less than 0.9. 
    Not only is our model stable unlike the coarse (Spectral 32) baseline, it also performs on par with a medium-grain (128) baseline, demonstrating a 4x improvement in spatial accuracy.
    If the $\physicsstep$ component of \Cref{eq:our_model} does not include the exponential filter as described in \Cref{sec:training_setup}, the $\learnedcorrection$ component diverges completely (\emph{Left} panel, ``No Filter'').
    }
    \label{fig:unstable_burgers}
\end{figure}

Burgers' equation is a simple one-dimensional non-linear PDE which is used as a toy model of compressible fluid dynamics.
The characteristic behavior of Burgers' equation is that it develops shock waves. Practitioners use this equation to test the accuracy of discretization schemes near discontinuities.

Like turbulent flows, the behavior of Burgers' equation is dominated by the convection, but unlike turbulent flows, it is not chaotic.
Prior studies of ML applied to Burgers' equation imposed random forcings~\citep{bar-sinai_learning_2019,um_solver---loop_2021}, but due to the non-choatic nature of the equation, discretization errors decay rather than compounding over time.
Instead, we use an unstable viscous Burgers' equation which slightly amplifies low frequencies in order to make the dynamics chaotic. \citet{sakaguchi_chaotic_1999} provides the following definition:
\begin{equation}\label{eq:unstable_burgers}
    \partial_t \vu = - u \partial_x \vu + \nu \partial_x^2 \vu + \int_0^L g(x - x')\partial_x^2 \vu(x') dx'
\end{equation}
for viscosity $\nu > 0$ and domain size $L$.
The three terms in this unstable Burgers' equation correspond to convection, diffusion and a scale-selective amplification of low-frequency signals.
The convolution $g(x - x')$ amplifies low frequencies, decaying smoothly to zero as the frequency increases but is best described in the Fourier space, which we defer to \Cref{app:unstable_burgers}.

Shock waves are challenging to model using the Fourier basis because there are discontinuities in the solution (\cref{sec:training_setup}). Due to the mixing of length scales introduced by convection, these errors can quickly propagate to affect the overall dynamics. This often results in instability at low resolutions, evidenced in our results (\cref{fig:unstable_burgers}, resolution 32). While there are classical methods for addressing discontinuities, e.g., adding a hyper-viscosity term, they often modify the underlying dynamics that they are trying to model. The inadequacy of unaltered spectral methods for solving this problem explains why our method is able to achieve approximately 4x improvement in spatial resolution over a spectral baseline. Results are summarized in \Cref{fig:unstable_burgers}. 

\subsection{2D turbulence}\label{sec:navier_stokes}

We consider 2D turbulence described by the incompressible Navier-Stokes equation with Kolmogorov forcing~\citep{kochkov_machine_2021}.
This equation can be written either in terms of a velocity vector field $\vv(x,y) = (\vv_x, \vv_y)$ or a scalar vorticity field \mbox{$\omega := \partial_x \vv_y - \partial_y \vv_x$}~\citep{boffetta_two-dimensional_2012}.
Here we use a vorticity formulation, which is most convenient for spectral methods and avoids the need to separately enforce the incompressibility condition~\mbox{$\nabla \cdot \vv = 0$}.
The equation is given by
\begin{equation}\label{eq:navier_stokes}
\partial_t\omega = - \vv\cdot\nabla \omega + \nu \nabla^2 \omega  - \alpha\,\omega + f,
\end{equation}
where the terms correspond to convection, diffusion, linear damping and a constant forcing $f$ (\cref{app:equation_params}).
The velocity vector field can be recovered from the vorticity field by solving a Poisson equation $-\nabla^2 \psi = \omega$ for the stream function $\psi$ and then using the relation $\vv(x,y) = (\partial_y \psi, -\partial_x \psi)$.
This velocity-solve operation is computed via element-wise multiplication and division in the Fourier basis.

For this example, we compared four types of models: spectral-only, FVM-only, EPD (ML-only), and two types of hybrid methods: our split operator method and a nonlinear term correction method.
The ML models are all trained to minimize the error of predicted velocities, which are obtained via a velocity-solve for the spectral-only, EPD and hybrid models that use vorticity as their state representation. The nonlinear term correction method uses a neural network to correct the inputs to the nonlinear term $\vv \cdot \nabla \omega$ and avoids using classical correction techniques, e.g., the so-called three-over-two rule \citep{orszag_elimination_1971}. \edit{Specifically, this approach takes as inputs $(\vv,\nabla\omega)$ and outputs a scalar field which is incorporating into the explicit part of the physics solver, e.g.\ $\partial_t\omega = - (\vv \cdot \nabla\omega + \operatorname{Nonlinear-Correction}(\vv, \omega;\theta)) + \eta\nabla^2 \omega - \alpha\omega + f$}

Results are summarized in \Cref{fig:navier_stokes}. Comparing Spectral 128 to FVM 1024 we can see that the spectral-only method, without any machine learning, already comes close to a similar improvement. Thus, a \textasciitilde 2x improvement over the spectral-only baseline --- achieved by both the EPD and our hybrid model --- is comparable to the \textasciitilde 8x improvement over FVM achieved in \citet{kochkov_machine_2021} in which they used a hybrid ML-Physics model.

\begin{figure}[t!]
    \includegraphics[width=1.0\textwidth]{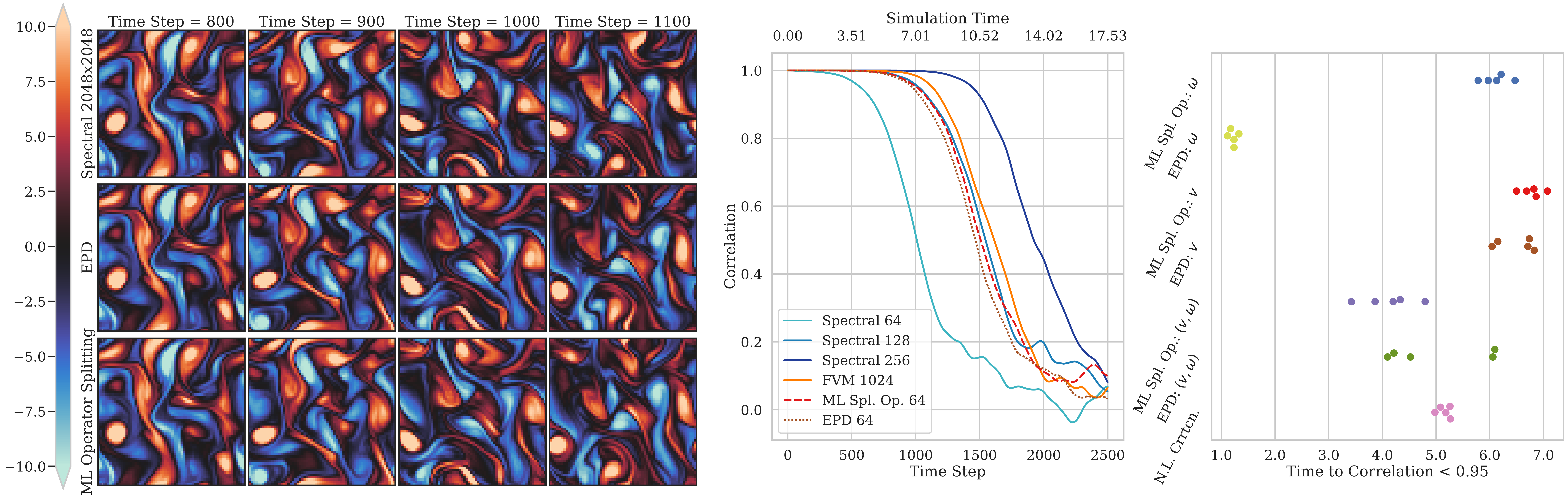}
    \caption{Benchmarking our \ourmodel (ML Spl.\ Op.) to spectral-only baselines and state-of-the-art ML-only, EPD model \citep{stachenfeld_learned_2022}. 
    \emph{Left}: Visualizing model states starting at the time when divergence begins to occur. At time step = 800, EPD, \ourmodel model, and the high-resolution spectral-only 1024x1024 correlate well. By time step = 900, both learned models have begun to decorrelate with the ground truth albeit in different ways.
    \emph{Right:} Comparing model variants across different random initializations. On the y-axis we measure the time to the first time step with correlation<0.95 with the high-resolution spectral ground truth. Then we compared our model with the EPD model across three data representation variations: velocity only ($\vv$), vorticity only ($\omega$) and velocity-vorticity concatenation~\mbox{($(\vv, \omega)$)} as well as with an Nonlinear Term Correction model (N.L.\ Crrtn.). Shown here are the best performing models over five different random neural network parameter initializations.
    }
    \label{fig:navier_stokes}
\end{figure}    

\paragraph{State representation.}
We compare three representations for inputs to neural networks in our EPD and split-operator methods: velocity, vorticity, and velocity-vorticity concatenation.
All models represent internal state with vorticity and learn corrections to vorticity, with the exception of velocity-only EPD models. These models never compute vorticity --- they use velocity to represent state as well as learned corrections, thus matching previous work by \citet{kochkov_machine_2021} and \citet{stachenfeld_learned_2022}.
Interestingly, this velocity-only representation has the best performance across the board, particularly for the fully learned EPD model, even outperforming velocity-vorticity concatenation (\cref{fig:navier_stokes}, right panel).

The nonlinear term~\mbox{$\vv\cdot\nabla \omega$} in \Cref{eq:navier_stokes} makes use of both vorticity and velocity representations. This indicates that in order to model the dynamics, the network must learn to solve for velocity --- a global operation --- making it challenging for a ConvNet restricted to local convolutions to learn.
In contrast, computing vorticity from velocity only requires evaluating derivatives, which can be easily evaluated with local convolutions, e.g., via finite differences.
This may explain the relatively-worse performance of Fourier Neural Operators using the velocity-based representation of \citet{stachenfeld_learned_2022} versus the vorticity-based representation of \citet{li_fourier_2021}.

\paragraph{Full versus nonlinear-term-only correction.}
To understand the value of time-splitting for the learned correction, we also performed an experiment where we instead incorporated our ML model into the nonlinear part of $\physicsstep$ (\cref{eq:our_model}). This entails solving the convection term with the same 4th order explicit Runge-Kutta method used in the numerical solver. Since velocity-space performed best with the split operator model, we also used it here.
The non-split learned correction has significantly worse performance.
Anecdotally, we observed that training with high order Runge-Kutta methods was significantly less stable. Models were much less robust to small changes in learning rates and dilation rates (\cref{app:architectures}) than models trained with first-order Runge-Kutta. We believe this may be due to the increased difficulty of propagating gradients through a network which is effectively four times deeper.
Another possibility is that the polynomial approximation of the high-order Runge-Kutta method introduces larger errors at the beginning of training when the model is less accurate.

\paragraph{Analyzing learning curves.}
To further compare the merits of different ML approaches, we compare the learning curves for select models in \Cref{fig:navier_stokes_learning_curve}.
First, we notice that although our best EPD and hybrid models are similarly accurate once trained, the hybrid models require drastically less compute to achieve fixed evaluation metrics.
For example, squared error of 2000 at 1024 time-steps requires only \num{3600} training steps for the median ML split operator velocity model, versus \num{164600} training steps for the EPD velocity model, corresponding to 4.4 versus 170 TPU-core hours.

These learning curves also reveal that our validation loss, in this case calculated over 32 unrolled time-steps, is not necessarily indicative of validation performance over much longer unrolls.
Although our models have never seen the exact validation data during training, many of them are still able to ``overfit'' to the task of predicting short trajectories, at the cost of generalization to long unrolls.
We also observe that different ML architectures overfit to different extents.
In particular, pure ML models and models with access to vorticity overfit more than our best hybrid model, the ML split operator with only velocity inputs.
The ML split operator with access to both velocity and vorticity is particularly interesting, because it seems to undergo a phase transition at around \num{30000} training steps, with correlation for long unrolls dropping dramatically as the model shifts into the ``memorization'' regime.

\section{Discussion}
\begin{figure}[t!]
    \centering
    \includegraphics[width=\textwidth]{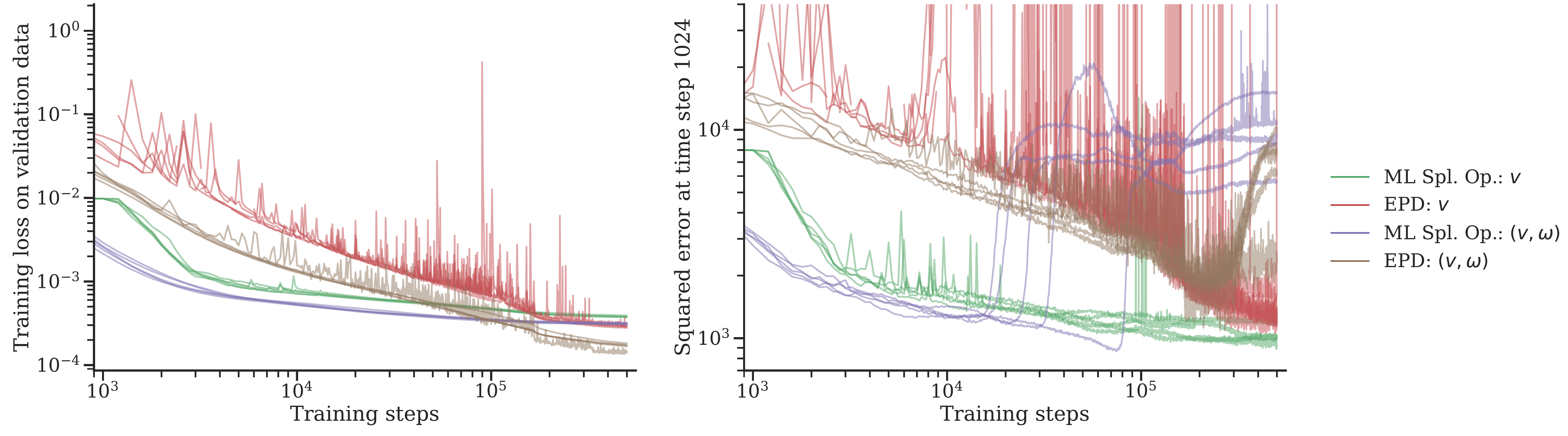}
    \caption{Learning curves for selected models, showing evaluation metrics after different numbers of training steps on the held-out validation dataset.
    The training loss is evaluated over the first 32 unrolled steps, whereas squared error is evaluated at 1024 steps ($t=7.2$). The latter better indicates the model performance we care about for predicting long trajectories.
    Separate lines show learning curves for each of our five randomly initialized training runs.
    }
    \label{fig:navier_stokes_learning_curve}
\end{figure}

In this paper, we demonstrated the potential of ML-augmented spectral solvers to improve upon the accuracy of spectral-only methods.
We also identified several key physically motivated modeling choices --- velocity-representation, first-order time stepping to improve sensitivity to hyperparameters, and the removal of global spatial artifacts --- which improve training for both ML-only and hybrid models.
Pure ML models can match the accuracy of hybrid models, but are considerably more expensive to train and show more indications of overfitting.

Traditional spectral methods are a powerful set of approaches for solving equations with smooth, periodic solutions.
It is yet unclear whether ML-based solvers can achieve meaningful computational speed-ups over classical spectral methods on PDEs of this class.
In contrast to prior work which showed computational speed-ups of up to 1-2 orders of magnitude over baseline finite volume~\citep{kochkov_machine_2021} and finite difference methods~\citep{list_learned_2022},
the roughly $\xspace 2 \times$ decrease in grid resolution for 2D turbulence with the ML split operator would allow for at most $8 \times$ reduction in computational cost.
However, our neural network for learned corrections is about $10 \times$ slower than $\physicsstep$, which counteracts this potential gain.
Small speed-ups might be obtained by using smaller networks or applying corrections less frequently, but overall there is little potential for accelerating smooth, periodic 2D turbulence beyond traditional spectral solvers.

Hybrid ML-spectral methods may enjoy more significant improvements on other problems.
Examples might include PDEs with less smooth solutions, such as 3D turbulence, where energy decays as $k^{-5/3}$ versus $k^{-3}$ in 2D~\citep{pope_turbulent_2000}. Or, global atmospheric models, where spectral methods do not achieve exponential convergence~\citep{Williamson2008-fvm-spectral}.
Cases where the exact governing equations are partially unknown, such as physical parameterizations for climate models~\citep{Brenowitz_nn_param_2018}, also present an opportunity for combining physical simulation with machine learning components.

 \newpage

\bibliographystyle{tmlr}
\bibliography{references_processed,references_revision}

\begin{thebibliography}{44}
\providecommand{\natexlab}[1]{#1}
\providecommand{\url}[1]{\texttt{#1}}
\expandafter\ifx\csname urlstyle\endcsname\relax
  \providecommand{\doi}[1]{doi: #1}\else
  \providecommand{\doi}{doi: \begingroup \urlstyle{rm}\Url}\fi

\bibitem[Bar-Sinai et~al.(2019)Bar-Sinai, Hoyer, Hickey, and
  Brenner]{bar-sinai_learning_2019}
Yohai Bar-Sinai, Stephan Hoyer, Jason Hickey, and Michael~P. Brenner.
\newblock \href{https://pnas.org/doi/full/10.1073/pnas.1814058116}{Learning
  data-driven discretizations for partial differential equations}.
\newblock \emph{Proceedings of the National Academy of Sciences}, 116, July
  2019.
\newblock ISSN 0027-8424, 1091-6490.
\newblock \doi{10.1073/pnas.1814058116}.

\bibitem[Boffetta \& Ecke(2012)Boffetta and
  Ecke]{boffetta_two-dimensional_2012}
Guido Boffetta and Robert~E. Ecke.
\newblock
  \href{https://www.annualreviews.org/doi/10.1146/annurev-fluid-120710-101240}{Two-{Dimensional}
  {Turbulence}}.
\newblock \emph{Annual Review of Fluid Mechanics}, 44, January 2012.
\newblock ISSN 0066-4189, 1545-4479.
\newblock \doi{10.1146/annurev-fluid-120710-101240}.

\bibitem[Boyd(1996)]{Boyd1996-wy}
John~P. Boyd.
\newblock \href{https://api.semanticscholar.org/CorpusID:17864531}{The
  {Erfc-Log} Filter and the Asymptotics of the Euler and Vandeven Sequence
  Accelerations}.
\newblock In {Ilin, A V and Scott, L Ridway} (ed.), \emph{Proceedings of the
  Third International Conference on Spectral and High Order Methods}, Houston,
  1996. Houston Journal of Mathematics.

\bibitem[Boyd(2001)]{Boyd:01_Chebyshev_Fourier}
John~P. Boyd.
\newblock
  \emph{\href{https://api.semanticscholar.org/CorpusID:119769668?utm_source=wikipedia}{{Chebyshev}
  and {Fourier} Spectral Methods}}.
\newblock Dover Books on Mathematics. Dover Publications, Mineola, NY, second
  edition, 2001.
\newblock ISBN 0486411834 9780486411835.

\bibitem[Brenowitz \& Bretherton(2018)Brenowitz and
  Bretherton]{Brenowitz_nn_param_2018}
N.~D. Brenowitz and C.~S. Bretherton.
\newblock
  \href{https://agupubs.onlinelibrary.wiley.com/doi/abs/10.1029/2018GL078510}{Prognostic
  Validation of a Neural Network Unified Physics Parameterization}.
\newblock \emph{Geophysical Research Letters}, 45, 2018.
\newblock \doi{https://doi.org/10.1029/2018GL078510}.

\bibitem[Bruno et~al.(2021)Bruno, Hesthaven, and
  Leibovici]{Bruno:2021_fc_dyn_nn_shocks}
Oscar~P. Bruno, Jan~S. Hesthaven, and Daniel~V. Leibovici.
\newblock \href{https://arxiv.org/abs/2111.01315}{{FC}-based shock-dynamics
  solver with neural-network localized artificial-viscosity assignment}, 2021.

\bibitem[Burns et~al.(2020)Burns, Vasil, Oishi, Lecoanet, and
  Brown]{Burns2020-gv}
Keaton~J Burns, Geoffrey~M Vasil, Jeffrey~S Oishi, Daniel Lecoanet, and
  Benjamin~P Brown.
\newblock
  \href{https://link.aps.org/doi/10.1103/PhysRevResearch.2.023068}{Dedalus: A
  flexible framework for numerical simulations with spectral methods}.
\newblock \emph{Phys. Rev. Research}, 2, April 2020.

\bibitem[Canuto et~al.(2007)Canuto, Hussaini, Quarteroni, and
  Zang]{canuto:2007spectral}
C.~Canuto, M.Y. Hussaini, A.~Quarteroni, and T.A. Zang.
\newblock \emph{\href{https://books.google.com/books?id=iDckv0W52cQC}{Spectral
  Methods: Evolution to Complex Geometries and Applications to Fluid
  Dynamics}}.
\newblock Scientific Computation. Springer Berlin Heidelberg, 2007.
\newblock ISBN 9783540307280.

\bibitem[Canuto et~al.(2006)Canuto, Yousuff~Hussaini, Quarteroni, and
  Zang]{Canuto2006-xw}
Claudio Canuto, M~Yousuff~Hussaini, Alfio Quarteroni, and Thomas~A Zang.
\newblock
  \emph{\href{http://link.springer.com/10.1007/978-3-540-30726-6}{Spectral
  methods: Fundamentals in single domains}}.
\newblock Scientific computation. Springer, Berlin, Germany, 1 edition, April
  2006.

\bibitem[Chen et~al.(2018)Chen, Rubanova, Bettencourt, and
  Duvenaud]{chen_neural_2019}
Ricky T.~Q. Chen, Yulia Rubanova, Jesse Bettencourt, and David~K Duvenaud.
\newblock
  \href{https://proceedings.neurips.cc/paper/2018/file/69386f6bb1dfed68692a24c8686939b9-Paper.pdf}{Neural
  Ordinary Differential Equations}.
\newblock In S.~Bengio, H.~Wallach, H.~Larochelle, K.~Grauman, N.~Cesa-Bianchi,
  and R.~Garnett (eds.), \emph{Advances in Neural Information Processing
  Systems}, volume~31. Curran Associates, Inc., 2018.

\bibitem[Cooley \& Tukey(1965)Cooley and Tukey]{Cooley_Tukey:1965}
J.~W. Cooley and J.~W. Tukey.
\newblock \href{http://www.jstor.org/stable/2003354}{An Algorithm for the
  Machine Calculation of Complex {F}ourier Series}.
\newblock \emph{Math. Comput.}, 19, 1965.
\newblock ISSN 00255718, 10886842.

\bibitem[Durbin(2018)]{Durbin2018-recent-developers-in-closure-modeling}
Paul~A Durbin.
\newblock \href{http://dx.doi.org/10.1146/annurev-fluid-122316-045020}{Some
  Recent Developments in Turbulence Closure Modeling}.
\newblock \emph{Annu. Rev. Fluid Mech.}, 50, 2018.

\bibitem[Fan et~al.(2019)Fan, Feliu-Fab{\`a}, Lin, Ying, and
  Zepeda-N{\'u}{\~{n}}ez]{FanYing:mnnh2019}
Yuwei Fan, Jordi Feliu-Fab{\`a}, Lin Lin, Lexing Ying, and Leonardo
  Zepeda-N{\'u}{\~{n}}ez.
\newblock \href{https://link.springer.com/article/10.1007/s40687-019-0183-3}{A
  multiscale neural network based on hierarchical nested bases}.
\newblock \emph{Research in the Mathematical Sciences}, 6, Mar. 2019.
\newblock ISSN 2197-9847.
\newblock \doi{10.1007/s40687-019-0183-3}.

\bibitem[Frezat et~al.(2022)Frezat, Le~Sommer, Fablet, Balarac, and
  Lguensat]{Frezat2022-fs}
Hugo Frezat, Julien Le~Sommer, Ronan Fablet, Guillaume Balarac, and Redouane
  Lguensat.
\newblock \href{http://arxiv.org/abs/2204.03911}{A posteriori learning for
  quasi-geostrophic turbulence parametrization}.
\newblock \emph{arXiv}, April 2022.

\bibitem[Gottlieb \& Hesthaven(2001)Gottlieb and
  Hesthaven]{gottlieb_spectral_2001}
D.~Gottlieb and J.~S. Hesthaven.
\newblock
  \href{https://www.sciencedirect.com/science/article/pii/S0377042700005100}{Spectral
  methods for hyperbolic problems}.
\newblock \emph{Journal of Computational and Applied Mathematics}, 128, March
  2001.
\newblock ISSN 0377-0427.
\newblock \doi{10.1016/S0377-0427(00)00510-0}.

\bibitem[Gottlieb \& Orszag(1977)Gottlieb and
  Orszag]{gottlied_orszag1977spectral}
David Gottlieb and Steven~A. Orszag.
\newblock
  \emph{\href{https://epubs.siam.org/doi/abs/10.1137/1.9781611970425}{Numerical
  Analysis of Spectral Methods}}.
\newblock Society for Industrial and Applied Mathematics, 1977.
\newblock \doi{10.1137/1.9781611970425}.

\bibitem[Huang et~al.(2022)Huang, Liang, Zhang, Yang, and
  Lin]{huang_accelerating_2022}
Zhongzhan Huang, Senwei Liang, Hong Zhang, Haizhao Yang, and Liang Lin.
\newblock \href{http://arxiv.org/abs/2208.03680}{Accelerating {Numerical}
  {Solvers} for {Large}-{Scale} {Simulation} of {Dynamical} {System} via
  {NeurVec}}, August 2022.
\newblock arXiv:2208.03680.

\bibitem[Kochkov et~al.(2021)Kochkov, Smith, Alieva, Wang, Brenner, and
  Hoyer]{kochkov_machine_2021}
Dmitrii Kochkov, Jamie~A Smith, Ayya Alieva, Qing Wang, Michael~P Brenner, and
  Stephan Hoyer.
\newblock \href{https://www.pnas.org/content/118/21/e2101784118}{Machine
  learning--accelerated computational fluid dynamics}.
\newblock \emph{Proc. Natl. Acad. Sci. U. S. A.}, 118, May 2021.

\bibitem[Kopriva(2009)]{Implementing_SM_PDEs:2009}
David~A. Kopriva.
\newblock \emph{\href{https://dl.acm.org/doi/10.5555/1643564}{Implementing
  Spectral Methods for Partial Differential Equations: Algorithms for
  Scientists and Engineers}}.
\newblock Springer Publishing Company, Incorporated, 1st edition, 2009.
\newblock ISBN 9048122600.

\bibitem[Li et~al.(2020)Li, Kovachki, Azizzadenesheli, Liu, Bhattacharya,
  Stuart, and Anandkumar]{graph_fmm:2020}
Zongyi Li, Nikola Kovachki, Kamyar Azizzadenesheli, Burigede Liu, Kaushik
  Bhattacharya, Andrew Stuart, and Anima Anandkumar.
\newblock \href{https://dl.acm.org/doi/abs/10.5555/3495724.3496291}{Multipole
  Graph Neural Operator for Parametric Partial Differential Equations}.
\newblock In \emph{Proceedings of the 34th International Conference on Neural
  Information Processing Systems}, NIPS'20, 2020.

\bibitem[Li et~al.(2021)Li, Kovachki, Azizzadenesheli, Liu, Bhattacharya,
  Stuart, and Anandkumar]{li_fourier_2021}
Zongyi Li, Nikola Kovachki, Kamyar Azizzadenesheli, Burigede Liu, Kaushik
  Bhattacharya, Andrew Stuart, and Anima Anandkumar.
\newblock \href{http://arxiv.org/abs/2010.08895}{Fourier {Neural} {Operator}
  for {Parametric} {Partial} {Differential} {Equations}}.
\newblock \emph{arXiv:2010.08895 [cs, math]}, May 2021.
\newblock arXiv: 2010.08895.

\bibitem[List et~al.(2022)List, Chen, and Thuerey]{list_learned_2022}
Björn List, Li-Wei Chen, and Nils Thuerey.
\newblock \href{http://arxiv.org/abs/2202.06988}{Learned {Turbulence}
  {Modelling} with {Differentiable} {Fluid} {Solvers}}.
\newblock \emph{arXiv:2202.06988 [physics]}, February 2022.
\newblock arXiv: 2202.06988.

\bibitem[McLachlan \& Quispel(2002)McLachlan and
  Quispel]{mclachlan_quispel_2002}
Robert~I. McLachlan and G.~Reinout~W. Quispel.
\newblock
  \href{https://www.cambridge.org/core/journals/acta-numerica/article/abs/splitting-methods/122F5736DAF3D88598989E68FE4D2EF2}{Splitting
  methods}.
\newblock \emph{Acta Numerica}, 11, 2002.
\newblock \doi{10.1017/S0962492902000053}.

\bibitem[Mishra(2018)]{mishra_machine_2019}
Siddhartha Mishra.
\newblock
  \href{https://www.aimspress.com/article/doi/10.3934/Mine.2018.1.118}{A
  machine learning framework for data driven acceleration of computations of
  differential equations}.
\newblock \emph{Mathematics in Engineering}, 1, 2018.
\newblock ISSN 2640-3501.
\newblock \doi{10.3934/Mine.2018.1.118}.

\bibitem[Orszag(1971)]{orszag_elimination_1971}
Steven~A. Orszag.
\newblock
  \href{http://journals.ametsoc.org/doi/10.1175/1520-0469(1971)028<1074:OTEOAI>2.0.CO;2}{On
  the {Elimination} of {Aliasing} in {Finite}-{Difference} {Schemes} by
  {Filtering} {High}-{Wavenumber} {Components}}.
\newblock \emph{Journal of the Atmospheric Sciences}, 28, September 1971.
\newblock ISSN 0022-4928, 1520-0469.
\newblock \doi{10.1175/1520-0469(1971)028<1074:OTEOAI>2.0.CO;2}.

\bibitem[Pope(2000)]{pope_turbulent_2000}
Stephen~B. Pope.
\newblock \emph{Turbulent {Flows}}.
\newblock Cambridge University Press, August 2000.
\newblock ISBN 9780521598866.

\bibitem[Pope(2004)]{pope_ten_2004}
Stephen~B Pope.
\newblock
  \href{https://iopscience.iop.org/article/10.1088/1367-2630/6/1/035}{Ten
  questions concerning the large-eddy simulation of turbulent flows}.
\newblock \emph{New Journal of Physics}, 6, March 2004.
\newblock ISSN 1367-2630.
\newblock \doi{10.1088/1367-2630/6/1/035}.

\bibitem[Roberts et~al.(2018)Roberts, Senan, Molteni, Boussetta, Mayer, and
  Keeley]{roberts2018climate}
Christopher~D Roberts, Retish Senan, Franco Molteni, Souhail Boussetta, Michael
  Mayer, and Sarah~PE Keeley.
\newblock \href{https://doi.org/10.5194/gmd-11-3681-2018}{Climate model
  configurations of the {ECMWF} Integrated Forecasting System ({ECMWF}-{IFS}
  cycle 43r1) for {HighResMIP}}.
\newblock \emph{Geoscientific model development}, 11, 2018.

\bibitem[Ronneberger et~al.(2015)Ronneberger, Fischer, and
  Brox]{ronneberger_u-net_2015}
Olaf Ronneberger, Philipp Fischer, and Thomas Brox.
\newblock
  \href{https://link.springer.com/chapter/10.1007/978-3-319-24574-4_28}{U-Net:
  Convolutional Networks for Biomedical Image Segmentation}.
\newblock In Nassir Navab, Joachim Hornegger, William~M. Wells, and
  Alejandro~F. Frangi (eds.), \emph{Medical Image Computing and
  Computer-Assisted Intervention -- MICCAI 2015}, Cham, 2015. Springer
  International Publishing.
\newblock ISBN 978-3-319-24574-4.

\bibitem[Sakaguchi(1999)]{sakaguchi_chaotic_1999}
Hidetsugu Sakaguchi.
\newblock
  \href{https://www.sciencedirect.com/science/article/pii/S0167278998003170}{Chaotic
  dynamics of an unstable {Burgers} equation}.
\newblock \emph{Physica D: Nonlinear Phenomena}, 129, May 1999.
\newblock ISSN 0167-2789.
\newblock \doi{10.1016/S0167-2789(98)00317-0}.

\bibitem[San \& Maulik(2018)San and Maulik]{san_extreme_2018}
Omer San and Romit Maulik.
\newblock \href{https://link.aps.org/doi/10.1103/PhysRevE.97.042322}{Extreme
  learning machine for reduced order modeling of turbulent geophysical flows}.
\newblock \emph{Physical Review E}, 97\penalty0 (4):\penalty0 042322, April
  2018.
\newblock \doi{10.1103/PhysRevE.97.042322}.
\newblock Publisher: American Physical Society.

\bibitem[Sanchez-Gonzalez et~al.(2020)Sanchez-Gonzalez, Godwin, Pfaff, Ying,
  Leskovec, and Battaglia]{sanchez-gonzalez_learning_2020}
Alvaro Sanchez-Gonzalez, Jonathan Godwin, Tobias Pfaff, Rex Ying, Jure
  Leskovec, and Peter~W. Battaglia.
\newblock \href{http://arxiv.org/abs/2002.09405}{Learning to {Simulate}
  {Complex} {Physics} with {Graph} {Networks}}.
\newblock \emph{arXiv:2002.09405 [physics, stat]}, September 2020.
\newblock arXiv: 2002.09405.

\bibitem[Stachenfeld et~al.(2022)Stachenfeld, Fielding, Kochkov, Cranmer,
  Pfaff, Godwin, Cui, Ho, Battaglia, and
  Sanchez-Gonzalez]{stachenfeld_learned_2022}
Kimberly Stachenfeld, Drummond~B. Fielding, Dmitrii Kochkov, Miles Cranmer,
  Tobias Pfaff, Jonathan Godwin, Can Cui, Shirley Ho, Peter Battaglia, and
  Alvaro Sanchez-Gonzalez.
\newblock \href{http://arxiv.org/abs/2112.15275}{Learned {Coarse} {Models} for
  {Efficient} {Turbulence} {Simulation}}.
\newblock \emph{arXiv:2112.15275 [physics]}, January 2022.
\newblock arXiv: 2112.15275.

\bibitem[Strang(1968)]{Strang:1968_operator_splitting}
Gilbert Strang.
\newblock \href{https://doi.org/10.1137/0705041}{On the Construction and
  Comparison of Difference Schemes}.
\newblock \emph{SIAM Journal on Numerical Analysis}, 5, 1968.
\newblock \doi{10.1137/0705041}.

\bibitem[Subel et~al.(2021)Subel, Chattopadhyay, Guan, and
  Hassanzadeh]{subel_data-driven_2021}
Adam Subel, Ashesh Chattopadhyay, Yifei Guan, and Pedram Hassanzadeh.
\newblock \href{http://arxiv.org/abs/2012.06664}{Data-driven subgrid-scale
  modeling of forced {Burgers} turbulence using deep learning with
  generalization to higher {Reynolds} numbers via transfer learning}.
\newblock \emph{Physics of Fluids}, 33\penalty0 (3):\penalty0 031702, March
  2021.
\newblock ISSN 1070-6631, 1089-7666.
\newblock \doi{10.1063/5.0040286}.
\newblock arXiv:2012.06664 [physics].

\bibitem[Tadmor(1986)]{Tadmor:1986_exp_acc}
Eitan Tadmor.
\newblock \href{https://doi.org/10.1137/0723001}{The Exponential Accuracy of
  {F}ourier and {C}hebyshev Differencing Methods}.
\newblock \emph{SIAM Journal on Numerical Analysis}, 23, 1986.
\newblock \doi{10.1137/0723001}.

\bibitem[Tran et~al.(2021)Tran, Mathews, Xie, and Ong]{tran_factorized_2021}
Alasdair Tran, Alexander Mathews, Lexing Xie, and Cheng~Soon Ong.
\newblock \href{http://arxiv.org/abs/2111.13802}{Factorized {Fourier} {Neural}
  {Operators}}.
\newblock \emph{arXiv:2111.13802 [cs]}, November 2021.
\newblock arXiv: 2111.13802.

\bibitem[Trefethen(2000)]{trefethen_spectral_2000}
Lloyd~N. Trefethen.
\newblock \emph{\href{https://doi.org/10.1137/1.9780898719598}{Spectral
  {Methods} in {MATLAB}}}.
\newblock Society for industrial and applied mathematics (SIAM), 2000.

\bibitem[Um et~al.(2020)Um, {Raymond}, {Fei}, Holl, Brand, and
  Thuerey]{um_solver---loop_2021}
Kiwon Um, {Raymond}, {Fei}, Philipp Holl, Robert Brand, and Nils Thuerey.
\newblock \href{http://arxiv.org/abs/2007.00016}{{Solver-in-the-Loop}: Learning
  from Differentiable Physics to Interact with Iterative {PDE-Solvers}}.
\newblock In \emph{{NeurIPS} 2020}, June 2020.

\bibitem[Wang et~al.(2020)Wang, Kashinath, Mustafa, Albert, and
  Yu]{wang_towards_2020}
Rui Wang, Karthik Kashinath, Mustafa Mustafa, Adrian Albert, and Rose Yu.
\newblock \href{http://dx.doi.org/10.1145/3394486.3403198}{Towards
  Physics-informed Deep Learning for Turbulent Flow Prediction}.
\newblock In \emph{Proceedings of the 26th {ACM} {SIGKDD} International
  Conference on Knowledge Discovery \& Data Mining}, 2020.

\bibitem[Williamson(2008)]{Williamson2008-fvm-spectral}
David~L Williamson.
\newblock \href{https://doi.org/10.1111/j.1600-0870.2008.00340.x}{Equivalent
  finite volume and Eulerian spectral transform horizontal resolutions
  established from aqua-planet simulations}.
\newblock \emph{Tellus Ser. A Dyn. Meteorol. Oceanogr.}, 60, January 2008.

\bibitem[Yeung \& Ravikumar(2020)Yeung and Ravikumar]{Yeung2020-3d-turbulence}
P~K Yeung and K~Ravikumar.
\newblock
  \href{https://link.aps.org/doi/10.1103/PhysRevFluids.5.110517}{Advancing
  understanding of turbulence through extreme-scale computation: Intermittency
  and simulations at large problem sizes}.
\newblock \emph{Phys. Rev. Fluids}, 5, November 2020.

\bibitem[Yin et~al.(2004)Yin, Clercx, and Montgomery]{Yin2004ParallelFourier}
Z.~Yin, H.J.H. Clercx, and D.C. Montgomery.
\newblock
  \href{https://www.sciencedirect.com/science/article/abs/pii/S004579300300077X}{An
  easily implemented task-based parallel scheme for the Fourier pseudospectral
  solver applied to 2D Navier–Stokes turbulence}.
\newblock \emph{Computers \& Fluids}, 33\penalty0 (4):\penalty0 509--520, 2004.
\newblock ISSN 0045-7930.
\newblock \doi{10.1016/j.compfluid.2003.06.003}.

\bibitem[Yu \& Koltun(2016)Yu and Koltun]{yu_multi-scale_2016}
Fisher Yu and Vladlen Koltun.
\newblock \href{http://arxiv.org/abs/1511.07122}{Multi-Scale Context
  Aggregation by Dilated Convolutions}.
\newblock In Yoshua Bengio and Yann LeCun (eds.), \emph{4th International
  Conference on Learning Representations, {ICLR} 2016, San Juan, Puerto Rico,
  May 2-4, 2016, Conference Track Proceedings}, 2016.

\end{thebibliography}

\newpage
\appendix

\section*{Appendix}

\section{Unstable Burgers}\label{app:unstable_burgers}

For convenience to the reader, we now provide the details of this equation as described in \citet{sakaguchi_chaotic_1999}. Writing \Cref{eq:unstable_burgers} in Fourier space yields a convenient representation. Consider the $k$-th wavenumber:
\begin{equation}
    \partial_t \uhat_k = -(\widehat g(k) + \nu) k^2 \uhat_k + \widehat\vf_k
\end{equation}
where $\widehat\vf_k$ represents the contribution of the nonlinear term and $\widehat g$ is the Fourier transform of the convolution $g$. Now we can simply let $\widehat g(k) = -.04 e^{-16 k^2}$. (Note: In \citet{sakaguchi_chaotic_1999}, the authors simply define $g$ is Fourier space).

As shown in \Cref{fig:g_fn}, this definition for $g$ amplifies low wave numbers while damping high wave numbers allowing for more complex dynamics than the original Burgers' Equation.

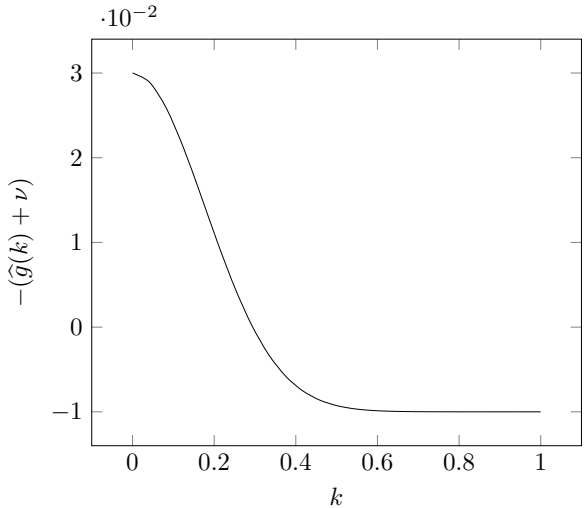
\begin{figure}[th!]
  \centering
\begin{tikzpicture}[smooth, domain=0:1, scale=.95]
  \begin{axis}
    [xlabel = $k$,
     ylabel = $-(\widehat g(k) + \nu)$]
    \addplot[mark=none]{-0.01 + 0.04 * e^(-16*x^2)};
  \end{axis}
\end{tikzpicture}
\caption{Plot of the scaling factor of the diffusion term in the Unstable Bergers' equation for each wave number $k$.}\label{fig:g_fn}
\end{figure}

\section{Data generation}\label{app:equation_params}

All training data was generated using pseudospectral solvers. We split the linear and nonlinear terms of the equation into implicit and explicit terms respectively, and used a Crank-Nicolson time stepping scheme with low storage fourth order Runge-Kutta, as described in Appendix D.3 of \citet{canuto:2007spectral}. Time step sizes were chosen according to the Courant–Friedrichs–Lewy (CFD) condition on the simulation grid, and are proportional increased for coarser simulations.
Downsampling was then performed in space (\cref{sec:training_setup}) and in time.
High resolution trajectories were downsampled to the target resolution (e.g.\ $N=32$ or $N=64$ grid points) using a block filter in Fourier space. Gibbs Phenomena were then removed from the downsampled trajectories using a low-pass filter as described in \Cref{sec:training_setup}.

For simplicity, in \Cref{sec:fourier_basis} we assumed that the spatial domain of $\vu$ is $[0, 2\pi]$. It is often convenient to regard the KS Equation and unstable Burgers' as having a larger domain since the domain size dictates the degree of chaos.

We generated data according to the following parameters:

\fbox{
\begin{minipage}{\textwidth}
\textbf{Kuramoto-Sivashinsky}
\begin{itemize}
    \item Spatial domain: $[0, 64]$, to match \citet{bar-sinai_learning_2019}
    \item Number of samples: 16
\item Warm-up time: \num{800.0}
    \item Simulation time: \num{800.0}
    \item Reference simulation grid: 1024
    \item Reference simulation time step: \num{0.0208333}
    \item Viscosity: $\nu = 0.01$
\end{itemize}
\end{minipage}
}

\fbox{
\begin{minipage}{\textwidth}
    \textbf{Unstable Burgers'}
    \begin{itemize}
        \item Spatial domain: $[0, 40\pi]$
        \item Number of samples: 16
        \item Warm-up time: \num{613.592}
        \item Simulation time: \num{50000}
        \item Reference simulation grid: 1024
        \item Reference simulation time step: \num{0.0613592}
        \item Viscosity: $\nu = 0.01$
    \end{itemize}
\end{minipage}
}

\fbox{
    \begin{minipage}{\textwidth}
        \textbf{2D Turbulence (Kolmogorov Flow)}
        \begin{itemize}
            \item Spatial domain: $[0, 2\pi] \times [0, 2\pi]$
            \item Number of samples: 16
            \item Warm-up time: 40.0
            \item Simulation time: 30.0
            \item Reference simulation grid: $2048 \times 2048$
            \item Reference simulation time step: \num{0.00021914011}
            \item Viscosity: $\nu = \SI{1.0e-3}{}$
            \item Drag coefficient: $\alpha = 0.1$
            \item Constant forcing $f((x,y)) = -k\cos(ky)$ with $k = 4$, which matches the velocity forcing from \cite{kochkov_machine_2021} acting on vorticity.
        \end{itemize}
    \end{minipage}
}

For each set of parameters, we created a training and a validation dataset, which only differ in the random seed used to generate the initial conditions.
All evaluation metrics are reported on the held-out validation data. \edit{By \emph{sample} we mean a sample from the initial conditions, thus 16 samples refers to a dataset which was constructed from 16 distinct initial conditions.}

\section{Learned model configurations}\label{app:architectures}

We followed the general architecture for Encoder-Process-Decoder models described in \citet{stachenfeld_learned_2022} but with different hyperparameters for 1D and 2D equations.

\paragraph{Encoder and Decoder modules.}
The Encoder and Decoders modules consisted of a single \edit{convolutional} layer with kernel size $= 5$. For 1D models we used 128 channels and for 2D models we used 64 channels.

\paragraph{Process module.}
The Process module consistent of residual blocks that uses dilated convolutions \citep{yu_multi-scale_2016}, interspersed with ReLU nonlinearities. Dilation rates are written as $(1, 2)$, which denotes a two-layer network whose first layer has a dilation rate of one (e.g.\ no dilation), and second layer has dilation rate of two.
For 1D models we used three residual blocks and for 2D Kolmogorov flow we used two blocks.

For 1D models the Process module had three layers with kernel size $ = 3$ and no dilation, i.e.\ with rates~$(1, 1, 1)$.

For Kolmogorov flow, we also used kernel size $ = 3$. We experimented with the following dilation rates: $(1,2)$ and $(1,2,4,2,1)$.
For each of these, we tried the following learning rates:
$[\scinot{1}{-5},
\scinot{5}{-4},
\scinot{2.5}{-4},
\scinot{1.0}{-4},
\scinot{5.0}{-3},
\scinot{1.0}{-3}]$.
While different models did seem to prefer certain dilation configurations, there was no general pattern across all models. For example, EPD with $(1,2)$ had the best performance whereas for our split operator method, $(1,2,4,2,1)$ had better performance.

\paragraph{Neural network parameter initialization.}
We also tested for the sensitivity to neural network parameter initialization shown in \Cref{fig:fvm_vs_spectral}, \emph{Right}. From the dilation configuration by learning rate sweep described above, we selected the best hyperparameter setting then tried 5 different random seeds for the neural network parameter initialization. Layer biases were initialized to zero and layer weights were initialized by truncated normal distributions.

\paragraph{Learning rates for 1D models.}
Generally, we found that performance was not significantly affected by learning rate, once a stable learning rate was found.

\paragraph{Number of unroll steps.}
The number of unroll steps for Kolmogorov flow was set to 32, and for both KS equation 8 and unstable Burgers' it was set to 8.

\paragraph{Optimizer.}
All machine learning models were trained using the Adam optimizer with $\beta_1 = 0.9$, $\beta_2 = 0.99$ and $\epsilon = \scinot{1}{-8}$. Our 2D turbulence models use a global batch size of 64 and train for $\scinot{5}{5}$ optimization steps. Each model takes approximately 65 wall-clock hours to train distributed across 8 TPU v4 cores.

\paragraph{Final model.}
Weights used for validation, including the learning curves plotted in \Cref{fig:navier_stokes_learning_curve}, use an exponential moving average of weights from training, with decay constant of 0.98.

\paragraph{Input/output scaling.}

Models which use $\ell_2$-loss generally make a normality assumption about error distribution. When this assumption is violated, the gradient of the parameters is often too large at the beginning of learning resulting in unstable trajectories which are impossible to recover from. To address this, we observed the scale of the errors with an untrained model and then rescaled the outputs of the $\learnedcorrection$ component so that the errors had variance one. Specifically, we used the following scales:
\begin{itemize}
    \item KS equation: 0.5
    \item Unstable Burgers': 0.1
    \item Kolmogorov Flow: 0.01
\end{itemize}
We found that for Kolmogorov Flow, it also helped to scale the inputs to the $\learnedcorrection$ term. We used a scale of 0.2 for all models on Kolmogorov Flow.

\edit{
\section{Background: Computational Fluid Dynamics}\label{sec:cfd}

\paragraph{Velocity-solve operation.}
By velocity-solve we mean a routine for mapping vorticity to velocity. Roughly speaking, this involves solving for the stream function and then uses the stream function to compute the velocity which is the standard approach. We do this in the frequency domain by inverting the Laplace operator and scaling by the frequency mesh in the x and y-directions appropriately. A quick sketch can be found in \citet{Yin2004ParallelFourier}.

\paragraph{Courant–Friedrichs–Lewy (CFL) condition.}
The CFL condition dictates the required time-step size given a spatial step-size to ensure stability of numerical solutions to PDEs. Informally, the time-step size cannot exceed the maximum rate of change (velocity) of the system. Thus,
\begin{equation}
\frac{\Delta x}{\Delta t}\leq C v_{\max}
\end{equation}
where $\Delta x$ is given by the desired spatial resolution, the maximum velocity $v_{\max}$ is computed empirically by running progressively finer simulation until convergence, and $C$ often depends on the type of time-stepping scheme.

\paragraph{Implicit and explicit time integration schemes.}
Explicit refers to the usual forward Euler integration scheme whereas implicit refers to the backward Euler integration scheme. The backward method, while more stable, often involves solving an equation algebraically to obtain an update step, e.g.\ $u_{t+1}$ in terms of $u_t$ -- thus the term implicit since it defines an update step only implicitly. Both methods are standard practice in any form of numerical simulation, and the choice of either type of method depends on the properties of the equation being solved.
}

\end{document}